\documentclass{article}

\usepackage{PRIMEarxiv}
\usepackage[utf8]{inputenc} 
\usepackage[T1]{fontenc}    
\usepackage{hyperref}       
\usepackage{url}            
\usepackage{booktabs}       
\usepackage{amsfonts}       
\usepackage{nicefrac}       
\usepackage{microtype}      
\usepackage{lipsum}
\usepackage{graphicx}

\usepackage{listings}
\usepackage{cite}
\usepackage{amsmath,amssymb}
\usepackage{algorithmic}
\usepackage{textcomp}
\usepackage{xcolor}
\usepackage{subcaption, wrapfig}
\usepackage{rotating}
\usepackage[font=small,skip=2pt]{caption}

\newcommand{\agg}[1]{{\color{black}#1}}
\newcommand{\ag}[1]{{\color{black}#1}}

\title{Explainable AI for Embedded Systems Design: A Case Study of Static Redundant NVM Memory Write Prediction
}

\author{
  Abdoulaye Gamati\'e, Yuyang Wang\\
  LIRMM, Univ. Montpellier - CNRS \\
  Montpellier, France\\
  \texttt{\{firstname.lastname\}@lirmm.fr} \\
}

\begin{document}
\maketitle

\begin{abstract}
This paper investigates the application of eXplainable Artificial Intelligence (XAI) in the design of embedded systems using machine learning (ML). As a case study, it addresses the challenging problem of static silent store prediction. This involves identifying redundant memory writes based only on static program features. Eliminating such stores enhances performance and energy efficiency by reducing memory access and bus traffic, especially in the presence of emerging non-volatile memory technologies. 

To achieve this, we propose a methodology consisting of: 1) the development of relevant ML models for \ag{explaining} silent store prediction, and 2) the application of XAI to explain these models. We employ \ag{two state-of-the-art} model-agnostic XAI methods to analyze the causes of silent stores. Through the case study, we evaluate the effectiveness of the methods. \agg{We find that these methods provide explanations for silent store predictions, which are consistent with known causes of silent store occurrences from previous studies. Typically, this allows us to confirm the prevalence of silent stores in operations that write the zero constant into memory, or the absence of silent stores in operations involving loop induction variables. This suggests the potential relevance of XAI in analyzing ML models' decision in embedded system design. From the case study, we share some valuable insights and pitfalls we encountered. More generally, this study aims to lay the groundwork for future research in the emerging field of XAI for embedded system design.}
\end{abstract}


\section{Introduction}
\label{sec:intro}

\ag{Machine learning (ML) is used in the design and optimization of embedded systems and Systems-on-Chip (SoCs) \cite{10317830, ajani2021overview, bhatt2020soc, rapp2021mlcad}.} Examples include identifying memory access patterns for prefetching \cite{osti_10187649} and optimizing branch prediction \cite{zangeneh2022using}. ML techniques often handle complex, nonlinear relationships within a system. Complex ML models are difficult to understand or explain. 

To tackle this challenge, \textit{eXplainable Artificial Intelligence} (XAI) \cite{xai:2021} has emerged, offering insights into model decision-making, and shedding light on potential biases and errors. The application of XAI to embedded system design aims to \textit{enable designers to optimize nonfunctional properties such as power and energy consumption, as well as interpret and enhance the functional behavior of the system}. 

Let us consider the design of an IoT device deployed in a smart home environment, where energy efficiency is critical. Data-driven modeling of the system design could be analyzed with XAI to examine how system components or functions contribute to overall energy consumption. Those insights could be used to redesign communication protocols or sensor sampling strategies during low activity periods. This would minimize energy consumption and reduce environmental impact. XAI-based iterative design of embedded systems for predicting equipment failures in industrial settings is another example. After initial deployment, users might find it challenging to trust the predictions due to occasional unexpected outcomes. XAI can help uncover that some sensor readings, while critical, are not well understood by users. On the basis of this insight, designers can iterate the model by adding new sensor data or adjusting features to better match users' interpretation of equipment health. 

\subsection{Considered problem} 
Nowadays, embedded system designs using ML models often ignore the rationale behind the model decisions. Addressing this typically relies on costly \textit{ad hoc} data analyses, as in \textit{static silent store prediction} \cite{pereira:2018}. This nontrivial problem focuses on improving program execution by predicting, at compile time (hence the term \textit{static}), which memory stores are likely to write the same value to a given location (hence the term \textit{silent}). 

Reducing silent stores helps mitigate the higher cost of writes compared to reads in emerging \textit{non-volatile memories} (NVMs), thus reducing latency and energy consumption \cite{pereira:2018, bouziane:2018}. During recent years, emerging NVMs have been extensively explored to improve the energy efficiency of embedded system architectures \cite{7309612, 7448986}. \agg{In terms of latency and energy consumption, NVM memory writes are known to be more costly than reads.} 
It also decreases write-backs and data bus traffic in multiprocessor systems \cite{silent-store:2000}. In the \textit{vortex} database benchmark from the SPEC CPU95 suite, up to 67\% of memory stores are silent, and potential write-back reductions can reach 81\% \cite{silent-store:2000}. 

Eliminating silent stores is based on the \textit{store-verify} method, where a \textit{store} operation is replaced by three steps: a \textit{load}, a \textit{comparison}, and the \textit{original store} itself. This process involves reading the target memory content, comparing it with the new value to be written; if they match, the store is skipped; otherwise, it is executed. \agg{Note that the store-verify method is similar to existing approaches such as the differential write method, which also suggests first reading the previous data in memory cells, comparing it with the new data, and only modifying cells whose values are different \cite{10.1145/1555754.1555759}. Using such a method, silent stores can be efficiently prevented at the hardware microarchitecture level. Nevertheless, it is not easily portable compared to software-oriented methods \cite{bouziane:2018, pereira:2018}, which integrate the store-verify optimization into program codes, ultimately making them executable on any hardware architecture.}

To minimize the performance overhead caused by \textit{store-verify}, it should only be applied to stores most likely to be silent. Hence, accurate silent store prediction is crucial, necessitating powerful ML models capable of capturing the intricate, nonlinear relationships between static program features and silent store occurrences, \ag{i.e., the \textit{possible causality between both}. In this work, we consider static silent store prediction to demonstrate how XAI can be used in embedded systems design.}

\ag{In this work, we tackle the challenge of building and explaining machine learning (ML) models for silent store prediction based on static program features.} We focus on the following questions: 

\begin{itemize}
\item \textit{\ag{(RQ1): How can we build relevant ML models for identifying silent store causes based on their predictions?} 
\item (RQ2): How can we effectively apply two state-of-the-art XAI methods for explaining these models?} 
\end{itemize}

\agg{To deal with the above questions, we rely on a previous study \cite{pereira:2018}, which already showed that static silent store prediction is beneficial to the energy efficiency of embedded systems integrating emerging non-volatile memory (NVM) technologies. To achieve this, the authors proposed experimental data to demonstrate the relevance of designing an ML-based prediction model to steer the effective application of the store-verify method. They trained a perceptron on a dataset. Afterward, they showed that depending on the benchmark, the energy cost of memory access can be improved up to a certain degree. In the present work, we use the same data set. However, we will focus on how to train effective machine learning models and how to interpret silent store predictions resulting from those models.} 

By addressing the aforementioned two questions, we identify some preliminary valuable insights into the application of XAI methods to static silent store prediction. \agg{Through this study, we also aim to illustrate how these methods could be adopted in ML-based embedded system design to explain model decisions.} 

\subsection{Our contribution} 
We propose a two-step methodology to answer both (RQ1) and (RQ2). 
\begin{itemize}
\item For (RQ1), we address a rationale for constructing suitable ML models tailored to silent store analysis. 
\item \ag{Regarding (RQ2), we define a sound and easily deployable approach for applying two XAI methods: \textit{SHapley Additive Explanation} (SHAP) \cite{shap:2017} and \textit{Anchors} \cite{anchor:Ribeiro_Singh_Guestrin_2018}.} 
\end{itemize}

To validate this methodology, we conduct a case study to explore static program features that might influence silent store predictions. \agg{Our preliminary findings suggest that the used XAI methods provide explanations for silent store predictions in accordance with previous studies that have identified possible causes of silent stores \cite{BellLL00, pereira:2018}. For instance, they establish the frequent occurrence of silent stores in operations that write the zero constant into memory, or the absence of silent stores in operations involving loop induction variables, i.e., variable increment with integer constants. 
This is a key step towards leveraging XAI methods and improving memory access costs}. This study not only confirms the effectiveness of these methods, but also uncovers potential pitfalls that can obscure XAI-based analysis results. \agg{Specifically, if the training dataset is unbalanced, precautions should be taken to ensure adequate prediction models are constructed for explainability.} Finally, we briefly discuss a few design targets related to embedded systems that could benefit from XAI.

\subsection{Outline}
\agg{
The rest of this document is organized as follows: Section \ref{sec:related} discusses some background notions and related work; then, Section \ref{sec:methodo} summarizes the reasoning methodology adopted in this work; Section \ref{sec:case_study} presents an application of the methodology in a case study to explain silent store prediction; Section \ref{sec:insight} discusses some general observations made during our study; finally, Section \ref{sec:conclu} gives some concluding remarks.
}

\section{Background notions and Related work}
\label{sec:related} 

XAI finds application in a wide range of tasks, including decision support, predictive maintenance, and anomaly detection \cite{apps:xai:app:2022}. Depending on the ML models used, explanations can be generated either early in the training process by using \textit{ante hoc} XAI methods or during inference by using \textit{post hoc} methods \cite{xai:2021}. \textit{Ante hoc} methods are suitable for transparent ML models (understandable by human experts), such as k-nearest neighbors, linear regression, and decision trees. On the other hand, \textit{post hoc} XAI methods rely on easily interpretable surrogate models that mimic complex base models with unknown inference mechanisms, like deep neural networks. These methods can be either \textit{model-agnostic}, applicable to any ML model, or \textit{model-specific}. Moreover, explanations can be either  \textit{local} or \textit{global}. A \textit{local} explanation aims to clarify why a model makes a prediction for a particular data instance. A \textit{global} explanation seeks to elucidate how the model generates predictions in general, drawing insights from its entire training dataset.

Despite its potential, there is very little (if any) application of XAI to the design of embedded systems \cite{xai-accel:2022, xai-others:2023}. The primary focus in \cite{xai-accel:2022} is on executing XAI on hardware accelerators, particularly to address transparency concerns with ML models. The concept of "explainable hardware" is introduced in \cite{xai-others:2023} as a means of achieving system-level explainability through XAI, although in the preliminary stages. Our approach aims to construct non-trivial ML models and explain their predictions based on static program features, addressing an original set of research questions and providing insights not yet covered.

Furthermore, existing research on silent store detection such as \cite{silent-store:2000,BellLL00,bouziane:2018,bouziane2017could} is predominantly based on dynamic analyses, involving the execution and profiling of computer programs. However, a distinctive approach, defined at compile-time, was introduced in \cite{pereira:2018}. It aims at predicting silent stores based on static program features. This is very challenging as silentness is identified based on the syntactic properties of a program. \ag{This approach does not address the explainability of the considered ML models and cannot explain the implicit reasoning behind the silentness prediction. In the present work, we deal with this issue by using model-agnostic \textit{post hoc} XAI methods. These explanations might later serve to avoid silent stores in program construction or compilation.}

\section{Reasoning methodology}
\label{sec:methodo}

We present a methodology for explaining silent store prediction. 

\vspace{-.2cm}
\subsection{Step 1: silent store classification}
\label{ssec:precision:recall}

We address a binary classification problem, i.e., predicting whether a store is silent or not. In the latter case, the store is referred to \ag{as} \textit{noisy}. 
We consider two common metrics\footnote{\agg{Note that other metrics could also be considered to evaluate adequate NN models. Indeed, beyond classification, the design of embedded systems can involve \textit{regression} or \textit{clustering} problems. For regression, \textit{Root Mean Squared Error} (RMSE), \textit{Mean Absolute Error} (MAE), and R2 score are the most commonly used metrics. The first two determine how accurate predicted values are, and their deviation threshold from actual values. The latter measures the percentage of correct predictions returned by a model. Clustering problems use \textit{similarity score} metrics, for example, the \textit{silhouette coefficient} that measures how well-defined and distinct clusters are.
}
} for evaluating binary classifiers: \textit{precision} and \textit{recall}. These are particularly important when dealing with imbalanced data sets, such as the one we borrowed from \cite{pereira:2018}. 

We denote \textit{true positive} (a silent store predicted as silent) and \textit{false positive} (a noisy store predicted as silent) by $TP$ and $FP$, respectively. Similarly, we denote \textit{true negative} (noisy store predicted as noisy) and \textit{false negative} (silent store predicted as noisy) by $TN$ and $FN$, respectively. \textit{Precision} quantifies how many of all predictions are true positives, while \textit{recall} measures how many actual silent stores are correctly predicted, i.e.: $precision = TP/(TP+FP)$ and $recall = TP/(TP+FN)$. 
Increasing \textit{precision} by lowering $FP$ in its denominator will result in a decrease in \textit{recall} as $TP$ will grow by definition but not $FN$, and \textit{vice versa}. 
Then, how to decide a suitable balance between both metrics w.r.t. model explainability? 

Let us consider the issue of silent store elimination for NVMs \cite{pereira:2018}. NVMs are known for their significantly higher write costs, both in terms of latency and energy, compared with read costs \cite{bouziane:2018}. Upon a false positive prediction, applying the \textit{store-verify} method imposes an unnecessary \textit{read-and-compare} execution overhead. Assuming the \textit{compare} operation is negligible compared with reading, let us denote the cost of a read by $C_r$. Upon a false negative prediction, an opportunity to replace a store with a more efficient \textit{read-and-compare} operation is missed. In essence, if $C_s$ represents the cost of a store, failing to make this replacement results in missing a potential savings of $C_s - C_r$. Now, let us assume a prediction model is prone to making a fixed number $m$ of mistaken predictions, i.e., $FP + FN$. 
If this model predicts a percentage $p$ of false positives, compared to a perfect predictor, it incurs overhead as follows:

\begin{equation}\label{eq:cost-fp}
m \times p \times C_r
\end{equation}

Likewise, the proportion of $FN$ for the same model will incur the following overhead compared with a perfect predictor: 

\begin{equation}\label{eq:cost-fn}
 m \times (1-p) \times (C_s - C_r)
\end{equation}

The total overhead of the model compared with a perfect predictor is defined by the sum of (\ref{eq:cost-fp}) and (\ref{eq:cost-fn}), as follows:

\begin{equation}\label{eq:cost-total}
m \times (C_s-C_r)-m \times (C_s - 2 \times C_r) \times p
\end{equation}

If $C_s > (2 \times C_r)$, increasing $p$ in expression (\ref{eq:cost-total}) will result in lower overhead. According to NVM technology designs, $C_s$ can be $1.02$ to $75$ times higher than $C_r$ \cite{bouziane:2018}. The corresponding prediction model will have lower precision and higher recall. It is suitable for implementing the \textit{store-verify} as shown in  \cite{pereira:2018}.

Conversely, when analyzing the significance of static features in relation to silent store prediction, it is advisable to employ more precise ML models that minimize the percentage $p$ of $FP$ as much as possible. Such models must have \textit{higher precision than recall}. 
As a result, in this work, we will train ML models taking this objective into account 
for explaining silent store predictions.

\subsection{Step 2: application of selected XAI methods} 
\label{ssec:algoXAI}

\ag{
Combining global and local model-agnostic XAI methods, we examine the impact of static program features on silent store prediction. Global explanations are validated using local methods.
In addition, we look for expressive and human-interpretable methods.

A fast global method we mentioned is \textit{Accumulated Local Effects} (ALE) \cite{ale:2020}, which measures the marginal impact of a single feature on predicted output. By varying the value of the feature while keeping the other features constant, perturbed data instances are generated. The differences in predictions between perturbed and original instances are computed to determine local effects of a feature. In static silent store prediction, feature interaction effects are crucial as they favor silentness \cite{pereira:2018}. ALE is not well-suited to this. So, we use the \textit{SHapley Additive Explanation} (SHAP) method \cite{shap:2017}. This method allows ML models to be interpreted in a unified manner, accounting for isolated feature effects and feature interactions. SHAP is more computationally expensive than ALE.

To validate SHAP explanations, we exploit a fast local explanation method. The very popular \textit{Local Interpretable Model-agnostic Explanations} (LIME) method \cite{lime:ribeiro2016} and its CLIME variant \cite{clime:2022} are typical candidates. LIME explains models by learning a linear decision boundary that approximates a model w.r.t. a perturbation space. In CLIME, the user can specify Boolean constraints on this space to generate more focused perturbations. 
When applied to the static silent problem, these methods provide similar explanations for supposedly significant and less significant features. Consequently, it is difficult to discern the real impact of the different features. This limitation is possibly due to the way the methods handle feature space perturbation. 
We select the Anchors method, which explains non-linear boundaries in ML model behavior \cite{anchor:Ribeiro_Singh_Guestrin_2018}.  
}

\subsubsection{SHAP method}
\label{sec:methodo:SHAP}
Let us consider a reference set $\Phi$ of features. SHAP computes the \textit{Shapley value} of a feature $\phi_i \in \Phi$, w.r.t. a \textit{feature vector} $\phi=\langle \phi_1, ..., \phi_k\rangle$ and a function $f$ corresponding to a prediction model. This value is the average contribution of feature $\phi_i$ to predictions by $f$ w.r.t. $\phi$, defined by function $\Psi_i$ \cite{shap:2017}:

\begin{equation}\label{eq:shapley}
\Psi_i(f, \phi) = \Sigma_{z' \subseteq \phi'} 
\frac{|z'|!(|\Phi|-|z'|-1)!}{|\Phi|!}
[f_\phi(z')-f_\phi(z' \backslash \phi_i)]
\end{equation}
where $\phi'$ denotes a simplified interpretable binary vector that maps to the original input vector $\phi$ using a mapping function $h_\phi$. Typically, $\phi = h_\phi(\phi')$ transforms a binary input vector $\phi'$ into the original input feature space, where $\phi'_i = 1$ (or $0$) in the binary vector indicates the presence (or absence) of feature $\phi_i$ in the mapped original vector. The variable $z'$ belongs to the set $\{0, 1\}^{|\Phi|}$ where $|\Phi|$ is the number of features in $\Phi$. 
The notation $|z'|$ expresses the number of $1$'s in the binary vector $z'$; and $z' \subseteq \phi'$ represents the set of all vectors $z'$  where the $1$'s are a subset of those present in $\phi'$.
Here, $f_\phi(z') = f(h_\phi(z'))$ and the expression $[f_\phi(z')-f_\phi(z' \backslash \phi_i)]$ defines the marginal contribution of feature $\phi_i$ to the prediction function $f$ w.r.t. an input vector $z'$. Note that $z' \backslash \phi_i$ means the value at position $i$ in $z'$ is set to $0$. When evaluating $f$ on feature vectors with fewer elements than the original input vectors, the missing features are commonly assigned default values. In this case, we have chosen to use \textit{randomly chosen values} from the missing features within the working dataset $D$ as default values.  

In this study, each feature $\phi_i$ represents a specific static property of a store instruction \texttt{l: p[i]=v}, where \texttt{l}, \texttt{p}, \texttt{i} and \texttt{v} respectively denote the \textit{instruction label} in the program code, the \textit{pointer to the uploaded location}, the \textit{offset added to the pointer} when building the target store address, and the \textit{value to upload}. Table \ref{tab:features} summarizes the considered feature categories, based on \cite{pereira:2018}. 
Fig. \ref{fig:code:feature} sketches four static feature vectors associated with store instructions.

\definecolor{codegreen}{RGB}{50,205,50}
\definecolor{codegray}{RGB}{128,128,128}
\definecolor{codepurple}{RGB}{186,85,211}
\definecolor{backcolour}{RGB}{242,242,242}

\lstdefinestyle{mystyle}{
    backgroundcolor=\color{backcolour},   
    commentstyle=\color{blue}\bfseries,
    keywordstyle=\color{codepurple}\bfseries,
    numberstyle=\tiny\color{codegray},
    stringstyle=\color{blue},
    basicstyle=\scriptsize\ttfamily,
    breakatwhitespace=false,         
    breaklines=true,                 
    captionpos=b,                    
    keepspaces=true,                 
    numbers=left,                    
    numbersep=5pt,                  
    showspaces=false,                
    showstringspaces=false,
    showtabs=false,                  
    tabsize=4
}

\lstset{style=mystyle, language=C}

\begin{figure}[h]
\begin{lstlisting}
#define SIZE 5
int main() {
 int values[SIZE] = {1,2,3,4,5}; // (Vin, Sl0, Smn, Pay, Es1, ...)
 int sum = 0; // (Ozr, Vin, sz8, Sl0, Smn, ZER, Scm, ...)
 double average;
    
 for (int i = 0; i < SIZE; i++) 
    { sum += values[i]; } // (Vin, sz8, Sl1, Smn, Oic, ...)
 average = (double)sum / SIZE; // (Vdb, Sl0, Smn, Scm, DIV ...)
 ... }
\end{lstlisting}
   \caption{\ag{\small Static feature vectors specified within in C comments (lines 3, 4, 8 \& 9)}}
   \label{fig:code:feature}
\end{figure}

\begin{sidewaystable*}[htbp]
\centering
\caption{Static program feature categories, applicable to store instruction \texttt{l: p[i]=v}.}
\label{tab:features}
\begin{tabular}{|l||l|}
\hline
\textbf{Categories of features} &
  \textbf{Examples of features with their corresponding encoding between parenthesis} \\ \hline \hline
the type of variable \texttt{v} & floating point on 32 bits (\texttt{Vfp}), floating point on 64 bits (\texttt{Vdb}), integer (\texttt{Vin}), pointer (\texttt{Vpt}) \\ \hline
the size  in bits of the type associated with \texttt{v} &
  1 bit (\texttt{sz1}), 2 to 8 bits (\texttt{sz8}) \\ \hline
\begin{tabular}[c]{@{}l@{}} 
the program statements (or \textit{program slice}) that may\\ affect the values of \texttt{v} 
at some point of interest
\end{tabular} &
  \begin{tabular}[c]{@{}l@{}}addition operation (\texttt{ADD}), an LLVM get element pointer instruction (\texttt{GEP}), zero constant  (\texttt{ZER}),\\ integer constant (\texttt{INT}), subtraction operation  (\texttt{SUB}), division operation (\texttt{DIV}) \end{tabular} \\ \hline
the last instruction that produced the value of \texttt{v} &
  increment (\texttt{Oic}), assignment of zero (\texttt{Ozr}), integer constant update (\texttt{Oin}), load instruction (\texttt{Old}) \\ \hline
the type of pointer \texttt{p} &
  integer (\texttt{Pin}), C-like struct (\texttt{Pst}), array (\texttt{Pay}), SIMD vector (\texttt{Pvc}) \\ \hline
the location of the region pointed by \texttt{p} &
  static memory (\texttt{Msc}), stack (\texttt{Msk}), heap (\texttt{Mhp})  \\ \hline
the label \texttt{l} of the store instruction in a program &
  \begin{tabular}[c]{@{}l@{}}within the main function (\texttt{Smn}), outside a loop (\texttt{Sl0}), within a singly, doubly, or triply nested \\ loop (\texttt{Sl1, Sl2, Sl3}), compulsory store instruction (\texttt{Scm})\end{tabular} \\ \hline
\begin{tabular}[c]{@{}l@{}}the offset \texttt{i} that is added to \texttt{p} when \\ building the store addresses, i.e., \texttt{p+i}\end{tabular} &
  \begin{tabular}[c]{@{}l@{}} \texttt{i} has a stride of size 1 (\texttt{Es1}), or 8 (\texttt{Es8}), \texttt{i} is created by some affine expression, \\e.g., \texttt{i = 2*b + c} (\texttt{Eaf}) \end{tabular} \\ \hline
\end{tabular}
\end{sidewaystable*}

All the features considered in this study are binary in nature: $1$ means a specific feature is satisfied by a store instruction, while $0$ means it is not. 
In general, as previously observed in \cite{pereira:2018}, static silent stores are characterized by specific combinations of features, as illustrated in Fig. \ref{fig:code:feature}. Noteworthy feature combinations include:
\begin{itemize}
    \item 
    \textit{nullifier} due to null value updates, contains 
    \texttt{Ozr} or \texttt{ZER}; and
    \item 
    \textit{static memory initializer}, due to null value assignment in static memory, combines \textit{nullifiers} with \texttt{Msc}. 
\end{itemize}

As for noisy stores, notable feature combinations include
\begin{itemize}
    \item 
    \textit{induction} due to variable increment with integer constants, which often involves \texttt{ADD}, \texttt{Oin}, \texttt{Oic} and \texttt{INT}.
\end{itemize}

\ag{In SHAP, dependence plots are provided that display the relationship between two different features based on the model's predictions. Nonetheless, this is not sufficient to address the combined effects of more than two static program features on store silentness. To overcome this limitation,
we compute the \textit{combined SHAP value} for vectors $\phi^k$ of $k$ features, by proceeding as follows:} 1) we define a set of $\phi^k$ to analyze; 2) for each such $\phi^k$, we extract from the working dataset $D$, all static store data instances $d_j$ containing all features of $\phi^k$; 3) finally, for each $d_j$, we average the sum of the SHAP values of its features to the silent store prediction. 
 
More formally, the above algorithm can be defined as follows, with $f$ representing the silent store prediction model:

\begin{enumerate}
    \item define a set $\{\phi^k = \langle \phi_1, ..., \phi_k \rangle, k \geq 2\}$ of $k$-feature vectors, containing feature combinations of interest; there are $C_{n}^{k}$ possible combinations for each $k$, where $n=|\Phi|$;
    \item for each vector $\phi^k$, extract from the working dataset $D$ all sets $D_{|\phi^k}$ of $n$-feature vectors containing the components of $\phi^k$, as follows: $D_{|\phi^k}=\{d_{j, j \in 1..n} \in D,~s.t.~\phi^k \subseteq d_j\}$; 
    \item for each $D_{|\phi^k}$, average the sum of its feature components' SHAP values as follows:

\begin{equation}\label{eq:comb_shap}
\frac{[\Sigma_{j \in 1..n}\Psi_{1}(f,d_j)] + ... + [\Sigma_{j \in 1..n}\Psi_{k}(f,d_j)]}{n \times k}
\end{equation}
\end{enumerate}

To evaluate the \textit{soundness} of formula (\ref{eq:comb_shap}) w.r.t. combined feature importance, we also compute the exact ratio of $d_j$ \textit{instances actually labeled as silent} within $D_{|\phi^k}$, as follows: 

\begin{equation}\label{eq:silent-ratio}
\frac{|\{d_{j, j\in 1..n} \in D_{|\phi^k}~s.t.~label(d_j, D)=silent\}|}{|D_{|\phi^k}|}
\end{equation}

The validity of formula (\ref{eq:comb_shap}) is substantiated if the results obtained from formulas (\ref{eq:comb_shap}) and (\ref{eq:silent-ratio}) are correlated for each vector $\phi^k$, i.e.,  
vectors  exhibiting high (or low) combined SHAP values should correspond to high (or low) ratios of silent static store instances $d_j$. 

\subsubsection{Anchors method}
\label{sec:methodo:Anchor}
\ag{
Anchors \cite{anchor:Ribeiro_Singh_Guestrin_2018} is an XAI method that provides human-understandable local explanations for specific data instances. An explanation is a predicate (also referred to as "rule") that, when satisfied, results in the same output predicted by the trained model for a given data instance to be explained. 
}

In Anchors, a user may specify the portion of similar instances yielding the same prediction output, called \textit{precision}. When precision requirements need to be high, Anchors generates highly restrictive predicates, which are typically satisfied by only a \textit{small subset} of the dataset, commonly referred to as \textit{coverage}. In particular, Anchors is less computationally intensive than SHAP.

\section{Case study}
\label{sec:case_study}

We summarize the model training process. Then, we apply our methodology to assess feature importance in silent store prediction. 

\subsection{Model training for silent store prediction}

We use the silent store dataset made available by \cite{pereira:2018} for our analysis. This dataset comprises a total of 89K static silent stores, each initially associated with 127 static program features. These stores were collected through the profiling of 222 programs. For each static store, the original dataset provides the proportion of silent \textit{dynamic stores}, i.e., the number of executed instances of this static store that were silent during data collection. We, therefore, label a static store as "silent" only if all its dynamic stores are silent. About 10\% of static stores are always silent. We reduce correlated features based on \textit{Pearson correlation score}.  
This leads to 76 out of the original 127 features.
Note that after cleaning the dataset, identical $76$-feature instances emerged with different silentness labels (even though their original $127$-feature versions did not). About 52\% of original static store instances fall into this category. 
For our experiments, we use \textit{Google Colab} as our computing platform. 

We trained a multi-layer neural network (NN) using the above dataset. 
The NN model consists of five dense layers spanning from the input layer to the output layer. These layers have 64, 32, 16, 8, and 1 neurons, respectively. The first four layers use a \textit{relu} activation function, while the output layer uses \textit{sigmoid}. The dataset is split into 80\% and 20\% respectively for training and testing. \agg{For simplicity, we deliberately omitted a validation data subset. Here, having such a subset does not change our ML model's performance.} The learning rate is $3 e^{-4}$. The selected optimizer is \textit{Adam} 
We define 3000 training epochs (with \textit{early stopping}) and a batch size of 20,000. The best precision and recall values obtained are about 0.60 and 0.29 respectively. These hyperparameters provide us with the best learning outcome among all the evaluated hyperparameter configurations. 

\subsection{Feature importance analysis with SHAP}
In practice, the evaluation of SHAP values is computationally intensive. To mitigate this issue, we use an \textit{explainer} of SHAP Python library \cite{shaplib}, which approximates SHAP values for deep learning models, instead of the SHAP \textit{exact explainer}. 

\subsubsection{Single feature perspective}
\begin{figure}[htbp]
 \centering
\includegraphics[width=0.5\textwidth]{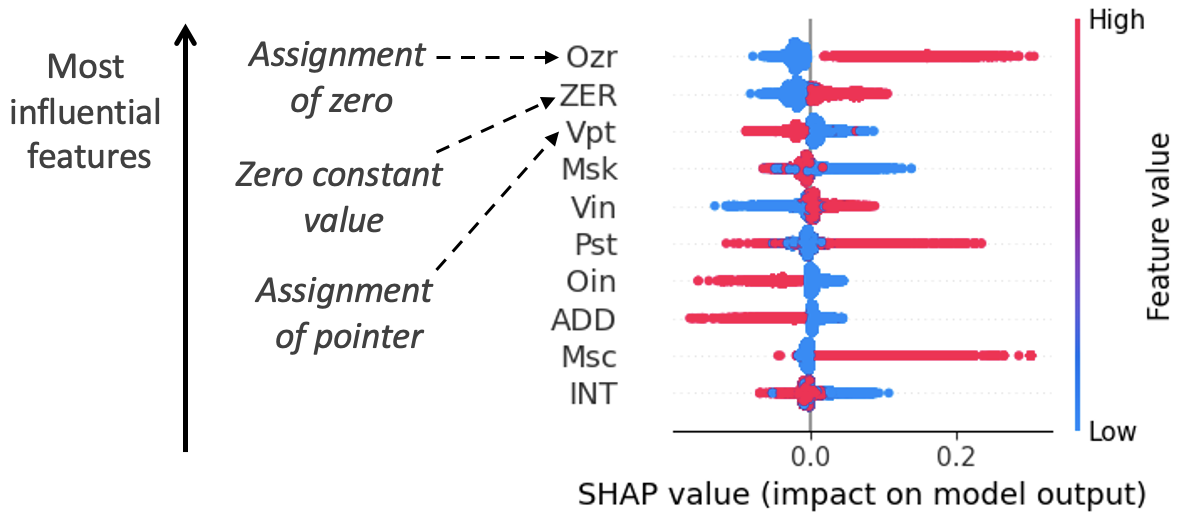}
    \caption{\small SHAP beeswarm plot over 10 most significant features w.r.t. silentness prediction by the trained NN model}
    \label{fig:shap10}
\end{figure}

Fig. \ref{fig:shap10} shows a SHAP beeswarm plot defined over 10 most influential features, \agg{among the 76 features mentioned in Table \ref{tab:features}}. The importance of features is shown from top to bottom, ordered by mean absolute SHAP values. It provides valuable insight into the impact of these features on the output of trained NN. Each point in the plot represents a specific data instance. Red and blue colors respectively indicate whether a store data instance satisfies or not a feature. The horizontal axis shows the SHAP value range of the features. 

The \textit{nullifier} features \texttt{Ozr} (i.e., stored value produced by an \textit{assignment of zero}) and \texttt{ZER} (i.e., stored value is a \textit{constant zero}) have the highest impact. When these features are satisfied by data instances, denoted by the presence of red points in the beeswarm plot, their SHAP values tend to increase. This improves the chance of predicting silent stores. As observed in \cite{BellLL00}, programmers frequently store zero as an explicit value or as a null pointer in their code. On the other hand, the \textit{induction} features \texttt{Oin} (i.e., the last instruction that produces the stored value is an \textit{integer constant update}) and \texttt{ADD} (that is, an \textit{addition operation} contributes to stored value) exhibit the opposite trend. When satisfied by stores, the SHAP values tend to decrease, possibly contributing to store noisiness. 
Unlike previous work \cite{BellLL00, pereira:2018}, \textit{loop-related features} indicating frequent store execution, do not highly contribute to store silentness in Fig. \ref{fig:shap10}.

\subsubsection{Combined feature perspective}
While Fig. \ref{fig:shap10} highlights the impact of isolated features, combined feature effects can further increase silentness prediction chances as discussed below. 
We apply formulas (\ref{eq:comb_shap}) and (\ref{eq:silent-ratio}) respectively to evaluate first the correlation between combined SHAP values and exact silentness ratios for various feature vectors. Then, we analyze feature combinations contributing the most to store silentness.

\begin{figure}[htbp] 
    \hspace{-2cm} 
    \centering 
    \begin{subfigure}
    {.18\textwidth}
     \includegraphics[width=5cm]{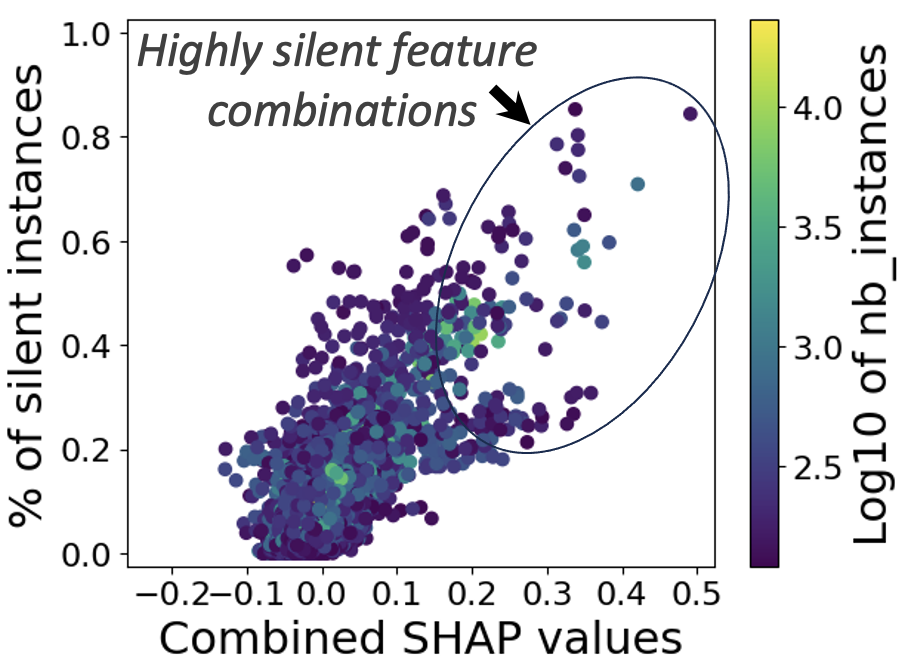}
       \caption{\small $\phi^3$ vectors}
    \label{fig:3features}
    \end{subfigure}
        \hspace{4cm} 
    \begin{subfigure}{.18\textwidth}
     \includegraphics[width=5cm]{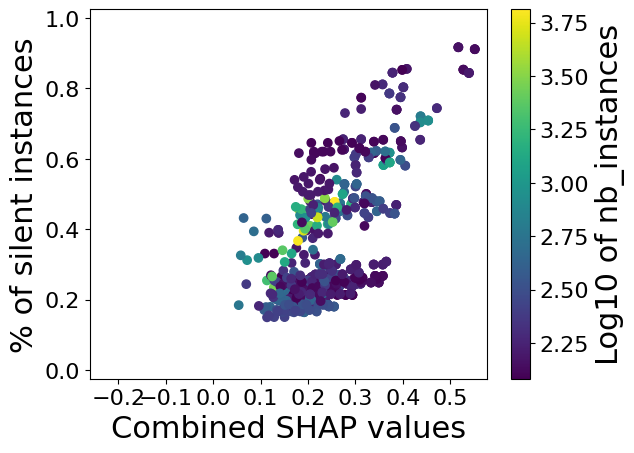}
       \caption{\small $\phi^4$ vectors}
    \label{fig:4features}
    \end{subfigure}
   \caption{\small Ratio of silent stores vs combined SHAP values over 3- and 4-feature vectors for NN (each point denotes a vector)}
   \label{fig:correlNN}
\end{figure}

Given that the complexity of static silent store analysis grows with the number of features to be mined per instruction, smaller feature vectors are more cost-effective than longer vectors.
In the sequel, we consider $\phi^k$ feature vectors with $k=3, 4$, which are relevant for illustration. 
There are $C_{76}^{3}$ = 70,300 possible $\phi^3$ vectors. Their enumeration on \textit{Google Colab} takes around 30 minutes. For each $\phi^3$, we compute formulas (\ref{eq:comb_shap}) and (\ref{eq:silent-ratio}). Enumerating all $\phi^4$ vectors is computationally expensive, so we define a subset of $\phi^4$ vectors by extending only $\phi^3$ vectors with a combined SHAP value greater than 0.2. Fig. \ref{fig:correlNN} summarizes the evaluation of our NN on three and four feature vectors respectively.

The trend depicted in Fig. \ref{fig:correlNN} reveals a reasonable correlation between the combined SHAP value (X-axis) and the exact silentness ratio (Y-axis) for $\phi^3$ and $\phi^4$ feature vectors. Each point on the plot denotes a $k$-feature vector. Its color represents the number (\textit{log.} scale) of static store instances satisfying the vector features. This number varies from 120 to 25000 in Fig.  \ref{fig:correlNN}. We deliberately omitted vector points  associated with less than 120 static store instances as they are less representative. 
The correlation exhibited in Fig. \ref{fig:correlNN} validates the soundness of combined SHAP values of features. 

\begin{figure}[htbp]
 \centering  \includegraphics[width=0.3\textwidth]{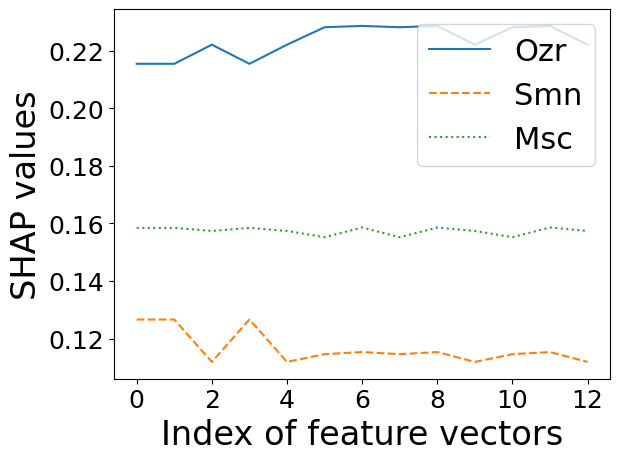}
    \caption{\small Feature contribution to silentness, over a few $\phi^4$ vector samples in which \texttt{Ozr}, \texttt{Smn} and \texttt{Msc} always occur
    }
    \label{fig:feature_contrib}
\end{figure}

We notice that feature vectors with high combined SHAP values, i.e., between 0.38 and 0.55, often include the features \texttt{Ozr}, \texttt{Smn}, and \texttt{Msc}. Their respective contributions to the combined SHAP values of twelve $\phi^4$ sample vectors are shown in Fig. \ref{fig:feature_contrib}. The \textit{nullifier} feature \texttt{Ozr} stands out as the major contributor with an individual SHAP value above 0.21. In contrast, the other two features, \texttt{Smn} and \texttt{Msc}, contribute less than 0.17 individually. Interestingly, in the absence of the latter two features, the SHAP value of \texttt{Ozr} drops to 0.13.

Beyond the above three features, we also observe separately the features \texttt{ZER}, \texttt{Scm}, and \texttt{Vin} in $\phi^4$ vectors that exhibit high combined SHAP values. The former has already been identified as an impactful \textit{nullifier} feature. As for the other two features, they respectively describe \textit{compulsory store instruction} in a program execution flow, and values \texttt{v} of \textit{integer type} in a store instruction \texttt{l: p[i]=v}. Integer type, due to its smaller value range, contributes more to silent stores than floating point type. This was previously shown in \cite{BellLL00}. This may explain the importance of \texttt{Vin} through its SHAP value.

For $\phi^4$ feature vectors with medium combined SHAP values, i.e., between 0.22 and 0.36, the feature \texttt{Sz1} emerges as an additional notable contributor to silent stores. It often characterizes Boolean value updates, favoring silent stores as observed in \cite{pereira:2018}.

\subsection{Validating SHAP explanations with Anchors}

Applying Anchors to the previous NN to generate local explanations requires a background dataset and the instance to explain. Here, we use the entire silent store dataset as the background dataset. The data instances to be explained are randomly chosen from those correctly predicted as silent. We impose that the generated "anchors" should have a precision of $0.95$ at least. 

The following predicate specifies an explanation generated by Anchors for a specific silent store instance: 

\begin{center}
(\texttt{Smn} $>$ 0.00) AND (\texttt{Ozr} $>$ 0.00) AND (\texttt{ZER} $>$ 0.00) AND (\texttt{Vin} $>$ 0.00) AND (\texttt{Scm} $>$ 0.00) AND (\texttt{INT} $\leq$ 0.00) 
\end{center}

\vspace{-.1cm}
This predicate essentially suggests static store instances satisfying 
(\texttt{Smn}=1, \texttt{Ozr}=1, \texttt{ZER}=1, \texttt{Vin}=1, \texttt{Scm}=1, \texttt{INT}=0),
are likely to be classified as silent by the NN model.  
It includes \textit{nullifier} features \texttt{Ozr} and \texttt{ZER}, combined with \texttt{Smn}, \texttt{Scm} and \texttt{Vin}. It denotes a null integer value assignment in a compulsory store executed in a main function. This is consistent with the results of the SHAP-based analysis. The combination of \textit{nullifier} features with \texttt{Smn} and \texttt{Scm} favors store silentness \cite{pereira:2018}.

\subsection{Leveraging the resulting explanations to avoid silent stores}

\agg{
We observed from the SHAP and Anchor explanations provided in the previous sections that certain features are helpful indicators of silent stores. It is the case, for example, with \textit{nullifier} features \texttt{Ozr} and \texttt{ZER}. Furthermore, we identified some feature combinations that are strongly associated with store silentness. In light of this knowledge, we can define a static program transformation that efficiently implements the store-verify method through a compiler. Given a program P, this process consists of the following steps: 

\begin{enumerate}
    \item \textit{analyze P and associate each store instruction in P with its corresponding static features;}
    \item \textit{apply the store-verify method to all store instructions with static features that strongly contribute to silent store occurrences.} 
\end{enumerate}

By only replacing store instructions with a high silentness probability according to the obtained explanations, the above program transformation reduces the potential overhead associated with the store-verify method. 
}

\section{Overall discussion}
\label{sec:insight}

We now discuss some insights and pitfalls from our study. \agg{The results of this study are useful for the future and the wider adoption of XAI in embedded system design.}

\subsection{Some insights and pitfalls}

\textit{Insight 1: Although our ML models were moderate but acceptable in accuracy, the results obtained with the selected XAI methods are meaningful in general.} We could not obtain very accurate ML models due to the inherent imbalance of the considered dataset. However, the derived explanations for static silent store causes were relevant and plausible in general. 

User expertise is sometimes necessary to spot potential inconsistencies. For example, the seemingly minor influence of loop-related features indicated by XAI in our study raises questions about the validity of this observation w.r.t. existing literature \cite{BellLL00, pereira:2018}.

\textit{Insight 2: Training alternative ML models instead of neural networks does not necessarily improve the explanation of silent stores.
}
Beyond NN, we also evaluated the Random Forest (RF) classification method, which is suitable for overcoming overfitting in decision tree training. RF harnesses the collective wisdom of weak decision tree learners to construct an accurate prediction model. \ag{Tree-based models inherently offer explainability \cite{burkart2021survey}, which makes RF an attractive candidate}.

\begin{figure}[htbp] 
    \hspace{-2cm}
    \centering
    \begin{subfigure}{0.18\textwidth}
    \includegraphics[width=5cm]{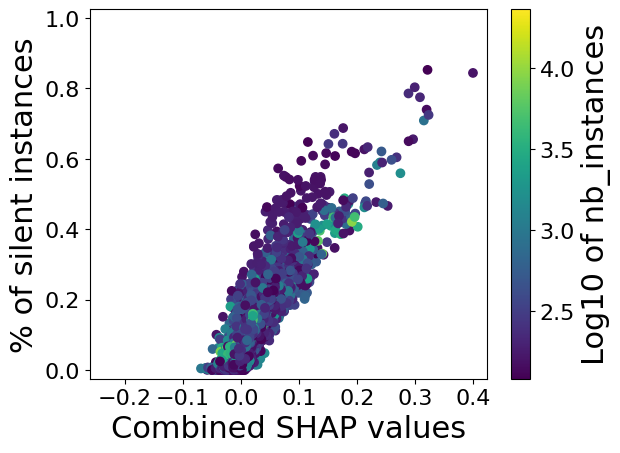}
        \caption{\small $\phi^3$ vectors}
    \label{fig:33features}
    \end{subfigure}
        \hspace{4cm}
    \begin{subfigure}{0.18\textwidth}
    \includegraphics[width=5cm]{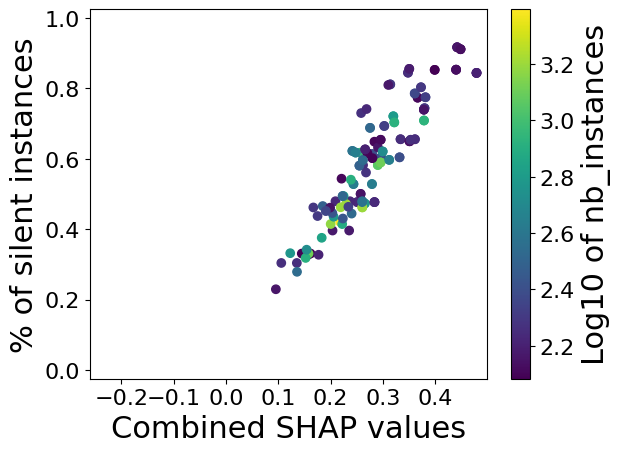}
        \caption{\small $\phi^4$ vectors}
\label{fig:44features}
   \end{subfigure}
   \caption{\small Ratio of silent stores vs combined SHAP values for RF} 
   \label{fig:correlRF}
\end{figure}

Consequently, we trained an RF model in which the resulting precision and recall values were $0.63$ and $0.23$, respectively (see Section \ref{ssec:precision:recall}). Fig. \ref{fig:correlRF} shows a correlation between the silentness ratio and the combined SHAP value for $\phi^3$ and $\phi^4$ vectors. 

The number of $\phi^4$ vectors obtained by extending $\phi^3$ vectors with combined SHAP values above 0.2, is fewer than that for NN (see Fig. \ref{fig:correlNN}). However, the features that contribute to storage silentness are similar between the RF model and the NN model. 

\begin{figure}[htbp]
    \hspace{-2cm} 
    \centering
    \begin{subfigure}
    {.18\textwidth}
     \includegraphics[width=5cm]{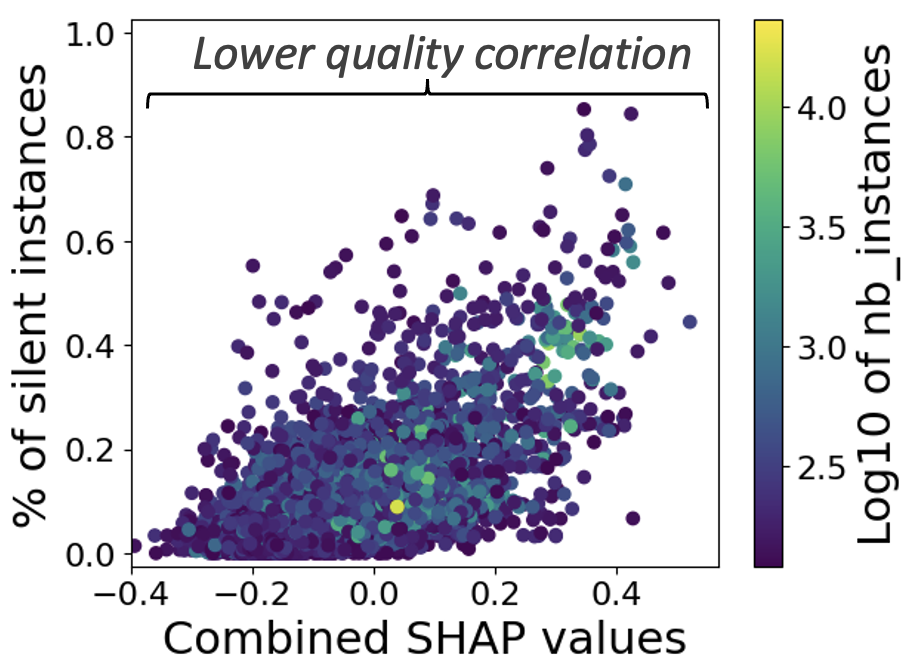}
       \caption{\small $\phi^3$ vectors}
    \label{fig:3featuresR}
    \end{subfigure}
        \hspace{4cm}
    \begin{subfigure}{.18\textwidth}
     \includegraphics[width=5cm]{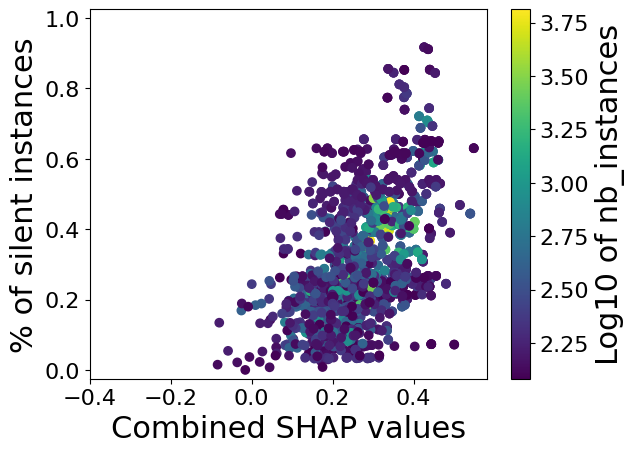}
       \caption{\small $\phi^4$ vectors}
    \label{fig:44featuresR}
    \end{subfigure}
   \caption{\small Ratio of silent stores vs combined SHAP values for NN, where precision is less than recall}
   \label{fig:correlR}
\end{figure}

\textit{Insight 3: Choosing the appropriate trade-off between precision and recall in ML model training is crucial for an acceptable application of XAI to classification tasks, in presence of unbalanced dataset.} Models with greater precision than recall are better suited for explainability given the unbalanced nature of our dataset (see Section \ref{ssec:precision:recall}). To support this claim, we evaluated the combined SHAP values for an NN with lower precision and greater recall. As depicted in Fig. \ref{fig:correlR}, more $\phi^3$ and $\phi^4$ are predicted to be silent by this NN due to higher recall. However, the correlation with the ratio of silent vector instances is not as strong as in Fig. \ref{fig:correlNN} due to a larger dispersion of vector points.  

Given this precision/recall trade-off issue in static silent store prediction, does it mean that ML model explainability and code optimization goals are not compatible? This is a compelling question that deserves further investigation. 
More generally, we believe that the suitable trade-off also depends on the requirements of target problems. When false positives are critical in classification, higher precision is preferable. However, when false negatives are more critical, a higher recall is the better option.

\textit{Pitfall 1: Low combined SHAP values should not be systematically  interpreted as noisy static store indications.
} Authors in \cite{pereira:2018} already identified the \textit{induction} features \texttt{Oic} and \texttt{ADD} as strong indicators of noisiness. This is confirmed in our study through their low SHAP values,  respectively $-0.04$ and $-0.03$ on average. However, this trend does not hold for combined SHAP values. Indeed, the working dataset $D$ used in our study is imbalanced, with the majority of static store instances being noisy. To effectively distinguish actual noisy stores from those incorrectly predicted as noisy, more accurate ML models are necessary. For instance, in Fig. \ref{fig:3features}, points with a negative combined SHAP values may have a silent store ratio between 0 and 0.6. This discrepancy likely results from inaccuracies in our NN, along with approximations in the employed SHAP \textit{explainer}.

\textit{Pitfall 2: The outcomes produced by the Anchors method are not always easily exploitable.
}
Beyond the above issue mostly related to SHAP explanations, Anchors predicates are sometimes needlessly complex, e.g., including several tens of irrelevant feature terms. From a human analysis standpoint, these lengthy predicates can make it challenging to extract the significance of highlighted feature properties. As an example, consider the following predicate of 31 feature terms, of which only two are satisfied, indicating the relevance of \textit{nullifiers}:

\begin{center}
(\texttt{Ozr} $>$ 0.00) AND (\texttt{ZER} $>$ 0.00) AND (\texttt{INT} $\leq$ 0.00) AND (\texttt{Vpt} $\leq$ 0.00) AND (\texttt{SUB} $\leq$ 0.00) AND ... 
\end{center}
\vspace{-.1cm}

Such complex explanations can lead to confusion rather than insight. Consequently, a method like Anchors should be carefully manipulated by users for effective analysis.

\agg{
\textit{Pitfall 3: Dataset quality concerning the store silentness problem affects the relevance of XAI-based explanations.  
}
Given the silent store data collection protocol used in \cite{pereira:2018}, it appears that several identical feature vectors can land in different classes, i.e., noisy and silent. Around 60\% of the dataset falls into this category. This means the dataset most likely does not accurately capture which features lead to silent or noisy stores. Among other factors, whether or not a store is silent also depends on the nature of the manipulated data. Consider the cumulative vector product as follows:}
\vspace{-.1cm}
\begin{verbatim}
void prod(int* v, int* w, int* x, const int N) {
 for (int i = 0; i < N; ++i)
   x += v * w;
}
\end{verbatim}
\vspace{-.1cm}

\agg{
If either v or w contains zeros, one will have lots of silent stores; otherwise, the stores will be noisy. So, the same program can have very different behaviors w.r.t. store silentness, depending on its input data. 

All features considered in \cite{pereira:2018} are syntactic, with the assumption that intuitively there is some relation between program syntax and semantics, although it is not very strong. For example, with the instruction \texttt{a = a + 1} the chance of a silent store is null. However, with the instruction \texttt{a = a + b*c}, if either \texttt{b} or \texttt{c} is zero, the store will be silent. Therefore, deciding whether the store is silent or not is not only affected by a syntactic analysis but also by a semantic analysis. 
}

\subsection{Towards XAI for embedded system design}

\agg{Embedded system devices have unique characteristics, including dedicated functionality and various resource and behavioral constraints: limited processing power, memory, energy, size, real-time response, etc. In many existing studies, ML has been applied alone at both design and device operation \cite{10317830, ajani2021overview}. For instance, ML models can be customized for performance, temperature, or power prediction, ensuring devices' constraints are met. They can also focus on anomaly detection, enabling fault identification during device operation, and facilitating predictive maintenance. To make this possible, one needs to collect suitable datasets and train ML algorithms accurately based on adequate input features, as discussed in the following examples. 

\begin{itemize}
    \item To predict execution time, such features may include software characteristics (e.g., code complexity and size), execution platform specifications (e.g., processor architecture, clock frequency, and cache sizes), and hardware-related execution events (e.g., cache misses, branch mispredictions and memory accesses) \cite{10.1007/s10617-023-09281-9}. For instance, code complexity can be measured using the diversity of instructions, functions or tasks \cite{GAMATIE20191}, while code size can be measured using the number of lines of code. Cache misses are a key source of execution time, while cache size is a critical factor in determining the number of cache misses during execution.  

    \item Power consumption prediction \cite{9142100} can be based on input characteristics such as system configurations (e.g., voltage levels and clock frequencies), system operating conditions (e.g., CPU load, memory usage and network activity), and system environmental factors (e.g., temperature and humidity). Higher voltage levels and clock frequencies, as well as heavier CPU loads, can typically lead to higher power consumption.

    \item Embedded system temperature prediction \cite{10.1007/978-3-031-49252-5_3} can be achieved using sensor data acquired within an embedded system, environmental conditions (e.g., ambient temperature and humidity), and system operational parameters (e.g., CPU usage and workload intensity). 

    \item Input features relevant to security prediction \cite{halak2022machine} include network data (e.g., incoming and outgoing traffic patterns and packet headers), system logs (e.g., authentication attempts and access patterns), and behavioral patterns (e.g., user interactions and any abnormalities within the system). For example, network data could be used to detect suspicious traffic patterns or authentication attempts in a system. Logs could help identify compromised user spaces or suspicious system activity. 

    \item Lastly, in the context of reliability prediction \cite{DJEDIDI2021114071}, relevant input features include failure data (e.g., previous instances of system failures or malfunctions), some testing data (e.g., data collected during stress tests or fault injection experiments), and environmental conditions (e.g., temperature, humidity and vibration levels during operation). Failure data may contain information about the failure type and when it happened. Testing data might indicate the performance scores of a system during tests and the number of conducted tests. 
\end{itemize}

The last two examples are potentially relevant to application fields including integrated decision-making processes for autonomous vehicles, medical diagnosis, or energy distribution in smart grids.

While the majority of existing work focuses on ML model precision, understanding the factors that mostly influence model decisions is not covered enough. Given the high diversity of input features considered in the above examples, it is very challenging to address this concern. XAI may fill this need by providing designers with a means to explain or interpret built ML models. It could contribute to make embedded systems design more transparent and understandable. Nevertheless, the successful combination of ML and XAI in embedded system design is not trivial. This is illustrated through the insights and pitfalls reported in our study, on the intricate problem of static silent store prediction. This preliminary study paves the way to a wider investigation of XAI for embedded system design.
}

\section{Concluding remarks}
\label{sec:conclu}

We considered XAI to study innovative approaches to embedded system design. This was illustrated through a pragmatic methodology applied to the static silent store prediction issue. Eliminating such redundant stores improves performance and energy efficiency. Leveraging cutting-edge XAI methods like SHAP and Anchors, we identified influential features from ML models to uncover some root causes of silent stores, aligning with prior research \cite{pereira:2018} and \cite{BellLL00}. For example, this allowed us to confirm the frequent occurrence of silent stores in operations that write the zero constant into memory, or the absence of them in operations involving loop induction variables.
The usage of explainable model predictions for silent store identification and elimination offers a new avenue for optimizing compiler techniques and hardware architectural design. 

While XAI holds promise, this preliminary study emphasized, however, the need for cautious application, highlighting both valuable lessons and potential pitfalls to know. Finally, it lays the foundation for future research in the emerging area of XAI for embedded system design. To consolidate our current insights, further ML problems, e.g., including regressions, must be investigated in the future.

\section*{Acknowledgements}
\agg{We thank Prof. Fernando Magno Quintao Pereira for many insightful comments and suggestions that contributed to improve this work.}

\bibliographystyle{unsrt}  
\bibliography{references}  

\end{document}